\pdfoutput=1

\documentclass[11pt]{article}

\usepackage[final]{acl}

\usepackage{times}
\usepackage{latexsym}
\usepackage{array}
\usepackage{tabularx}
\usepackage{graphicx}
\usepackage{xcolor}

\PassOptionsToPackage{skins,breakable}{tcolorbox}
\usepackage{tcolorbox}
\usepackage{algorithm}
\usepackage{algpseudocode}  
\usepackage{amsmath}
\usepackage{amssymb}
\usepackage{caption}    
\algrenewcommand\algorithmicindent{0.5em}
\algnewcommand{\LineComment}[1]{\State \(\triangleright\) #1}

\usepackage[T1]{fontenc}

\usepackage[utf8]{inputenc}

\usepackage{microtype}

\usepackage{inconsolata}

\usepackage{graphicx}

\usepackage{float}
\usepackage{booktabs} 
\usepackage{multirow}
\usepackage{makecell}
\usepackage{utfsym}
\usepackage{arydshln}
\usepackage{CJKutf8}

\usepackage[skins]{tcolorbox}
\newtcolorbox{pbox}[1][]{
colback=teal!15!white,
sharp corners,
#1
}

\usepackage{url}
\usepackage{breakurl}

\newcommand{\ours}{\textsc{MedPlan}}
\title{\ours: A Two-Stage RAG-Based System for Personalized Medical Plan Generation}

\author{
 \textbf{Hsin-Ling Hsu\textsuperscript{1}\thanks{Equal contribution}},
 \textbf{Cong-Tinh Dao\textsuperscript{2,3}\footnotemark[1]},
 \textbf{Luning Wang\textsuperscript{4}},
 \textbf{Zitao Shuai\textsuperscript{4}},
\\
 \textbf{Nguyen Minh Thao Phan\textsuperscript{2,3}},
 \textbf{Jun-En Ding \textsuperscript{5}},
 \textbf{Chun-Chieh Liao\textsuperscript{5}},
 \textbf{Pengfei Hu \textsuperscript{5}},
  \textbf{Xiaoxue Han \textsuperscript{5}},
\\
   \textbf{Chih-Ho Hsu \textsuperscript{6}},
   \textbf{Dongsheng Luo \textsuperscript{7}},
   \textbf{Wen-Chih Peng \textsuperscript{2}},
   \textbf{Feng Liu \textsuperscript{5}},
    \textbf{Fang-Ming Hung \textsuperscript{6}},
    \textbf{Chenwei Wu \textsuperscript{4}}
\\
 \textsuperscript{1}National Chengchi University,
 \textsuperscript{2}National Yang Ming Chiao Tung University,
\\
\textsuperscript{3}Can Tho University, 
\textsuperscript{4}University of Michigan,
 \textsuperscript{5}Stevens Institute of Technology,
\\
 \textsuperscript{6}Far Eastern Memorial Hospital
 \textsuperscript{7}Florida International University
\\
 \small{
   \textbf{Correspondence:} \href{mailto:chenweiwu99@gmail.com}{
chenweiwu99@gmail.com}
 }
}

\begin{document}
\maketitle
\begin{abstract}
Despite recent success in applying large language models (LLMs) to electronic health records (EHR), most systems focus primarily on assessment rather than treatment planning. We identify three critical limitations in current approaches: they generate treatment plans in a single pass rather than following the sequential reasoning process used by clinicians; they rarely incorporate patient-specific historical context; and they fail to effectively distinguish between subjective and objective clinical information.
Motivated by the SOAP methodology (Subjective, Objective, Assessment, Plan), we introduce \ours{}, a novel framework that structures LLM reasoning to align with real-life clinician workflows. Our approach employs a two-stage architecture that first generates a clinical assessment based on patient symptoms and objective data, then formulates a structured treatment plan informed by this assessment and enriched with patient-specific information through retrieval-augmented generation. Comprehensive evaluation demonstrates that our method significantly outperforms baseline approaches in both assessment accuracy and treatment plan quality. Our demo system and code are available at \href{https://github.com/JustinHsu1019/MedPlan}{https://github.com/JustinHsu1019/MedPlan}.

\end{abstract}

\section{Introduction}

\begin{figure}[htbp]
    \centering
    \includegraphics[width=0.5\textwidth]{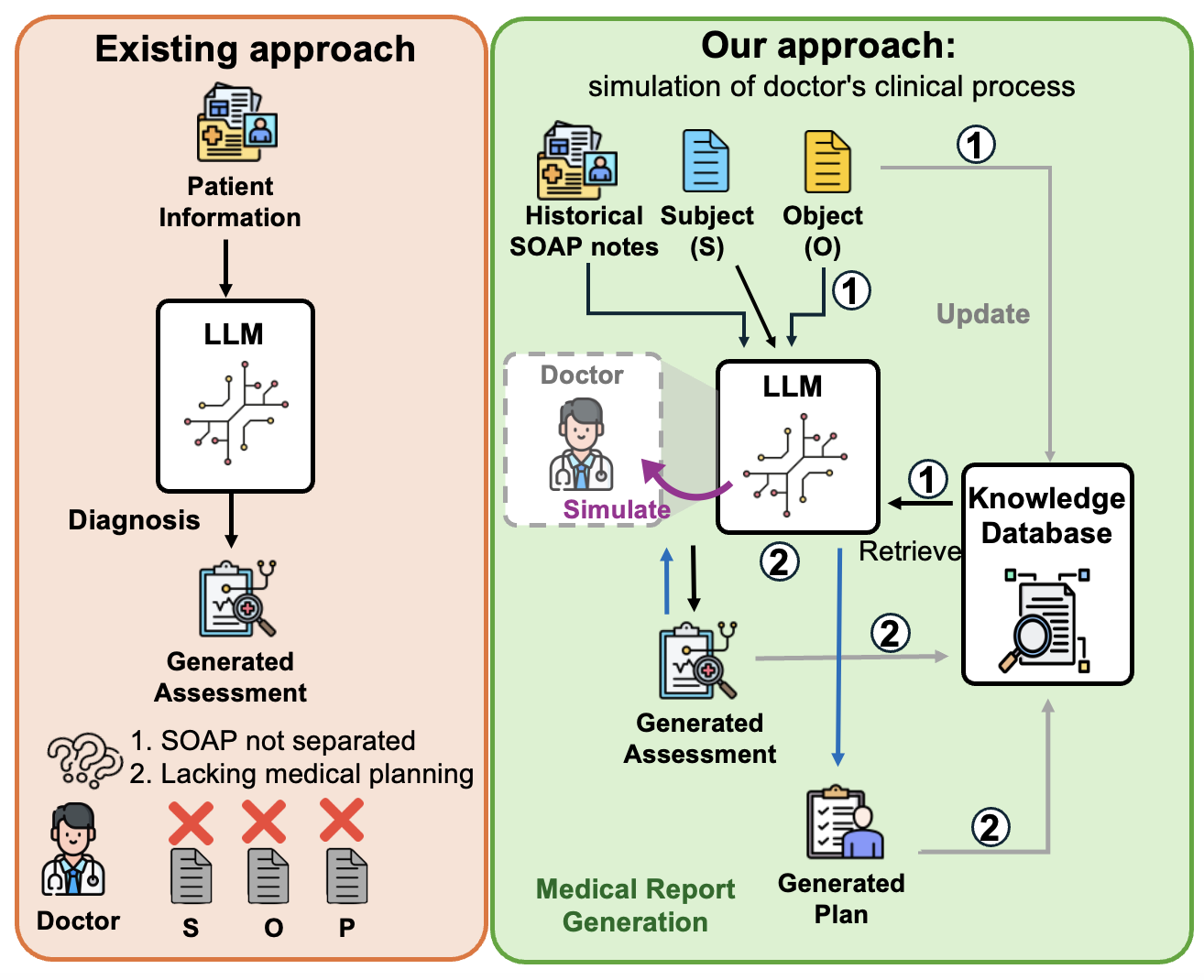}
    \caption{Compare the existing approach (left) with our proposed \ours{} (right). We adopt the SOAP protocol and simulate the doctor diagnosis process with LLM for medical plan generation.}
    \label{fig:intro}
\end{figure}

Deploying large language models (LLMs) for electronic health records (EHR)~\cite{evans2016electronic} analysis in high-stakes medical environments presents significant opportunities for enhancing patient care through automation and improved clinical decision support~\cite{yang2022large,zhang2024potential,sakai2025large,ding2024large}. Despite recent progress in adapting LLM to medical domain~\cite{tang2025medagentsbench, jiang2025medagentbench,restrepo2025multi}, most existing LLM systems~\cite{palepu2025towards,fan2024towards} for EHR focus largely on diagnostic assessment tasks, neglecting the crucial subsequent step of structured, patient-specific treatment planning~\cite{sarker2021defining,curtis2017translating}. Effective LLM-based planning could significantly reduce physician cognitive load, standardize care protocols, decrease treatment variability, and enable more personalized interventions.

Enabling LLM with trustworthy and personalized treatment planning capabilities introduces unique challenges—models must generate medically sound interventions, tailor recommendations to individual patient needs, and maintain a clear rationale connecting diagnosis to treatment~\cite{qiu2025quantifying}. Ideally, these systems should align with real-life clinical reasoning processes employed by healthcare professionals. The SOAP methodology (Subjective, Objective, Assessment, Plan) represents one of medicine's fundamental cognitive frameworks~\cite{sorgente2005soap,shechtman2002child}, systematically organizing clinical information into a structured sequential decision-making process. Under this protocol, clinicians first gather subjective patient-reported symptoms (S) and objective clinical data such as laboratory tests and physical examination findings (O). These elements provide the basis for a clinical assessment (A), subsequently informing a structured treatment plan (P).

However, our analysis identifies several critical limitations in current approaches. First, the few existing works on medical treatment planning with LLMs~\cite{liu2024automated, chen2025map} attempt to generate treatment plans directly from clinical data in a single pass, failing to mirror the sequential cognitive process physicians adopt, where clinicians first reach diagnostic conclusions before developing actionable interventions tailored to each patient's unique circumstances. This collapsed reasoning process risks producing treatment recommendations disconnected from their diagnostic foundations—a critical failure in medical decision-making where transparent causal relationships between findings and interventions are essential.

Second, current approaches rarely incorporate patient-specific historical context—such as medical history, previous treatment responses, and longitudinal trends—that physicians naturally consider when making treatment decisions. This neglect of personalized context leads to generic treatment recommendations that fail to account for individual patient factors crucial to treatment success. Finally, most systems don't effectively distinguish between subjective patient narratives and objective clinical measurements, despite this distinction being fundamental to clinical practice where a patient's subjective experience ("my chest hurts when I breathe") is weighed differently from objective findings (elevated troponin levels) in formulating both diagnoses and treatment plans.

These gaps motivate our research questions: \begin{itemize} \item \textbf{How can we structure LLM reasoning processes to mirror the sequential SOAP protocol used by clinicians, and does this improve treatment plan generation?} \item \textbf{How can we incorporate patient-specific contexts to better support individualized care decisions?} \end{itemize}

To address these challenges, we introduced \ours{}, a novel framework that explicitly structures LLM reasoning to mirror the SOAP clinical workflow. Our approach operates in two clinically-grounded stages that parallel physician cognitive processes: (1) a diagnostic phase where we generate an assessment (A) based on patient symptoms and clinical data (S and O), completing the diagnostic reasoning before proceeding, and (2) a therapeutic phase where we formulate a structured treatment plan (P) directly informed by the assessment and tailored to patient-specific factors. This two-stage architecture faithfully replicates how clinicians reason—first establishing what is happening before determining what should be done. We enhanced the planning phase through patient-specific retrieval-augmented generation (RAG)~\cite{lewis2020retrieval}, allowing the model to consider longitudinal patient information—mirroring how physicians integrate medical history into their treatment decisions.

Our contributions are three-fold: \begin{itemize} \item We introduced \ours{}, a novel SOAP-inspired two-stage LLM framework for EHR data that structures clinical reasoning to match physician workflows, providing reliable patient-specific assessments and plans. \item We conducted a comprehensive evaluation showing our method significantly outperformed baseline methods on various metrics in both clinical assessment and treatment plan generation. \item We released a fully functional system that tests our approach in a real clinical environment, allowing physicians to efficiently generate structured, patient-specific plans integrated with existing EHR workflows. \end{itemize}

\section{Related Work}

The SOAP framework has been widely recognized as a standard for clinical documentation and reasoning~\cite{cameron2002learning}. Several computational approaches have attempted to structure medical notes according to SOAP elements~\cite{castillo2019interaction}, but they typically treat these elements as documentation categories rather than as steps in a diagnosis reasoning process. Due to the success of LLMs, such as GPT-4, LLaMA, and Mistral-7B, these models have significantly impacted healthcare, particularly in medical documentation, clinical summarization, and decision support. Studies have demonstrated LLMs' potential in automating discharge note generation, extracting key clinical information from EHRs, and summarizing medical evidence, though challenges such as factual inconsistency and hallucinations remain~\cite{alkhalaf2024applying,tang2023evaluating}.

Recent research used patient physical information and examination results as input to make ChatGPT generate a series of initial diagnostic information, examination results, and recommended measures to create reports~\cite{zhou2023evaluation}. Additionally, RAG was used to improve the efficiency of medical document retrieval and integration of external knowledge~\cite{alkhalaf2024applying} or enhance the accuracy of LLMs in EHR summaries and medical note generation~\cite{yang2025retrieval}. However, current RAG applications primarily focus on data retrieval and aggregation without truly enhancing the internal generation process of LLMs, particularly when processing complex and large quantities of diagnostic reports to generate personalized diagnostic report plans. In this work, we provide a structured LLM retrieval process that incorporates multiple clinical text information while addressing past patient historical records using a two-stage pipeline for medical planning generation. 

\section{Methodology}

\begin{figure*}[htbp]
    \centering
    \includegraphics[width=1\textwidth]{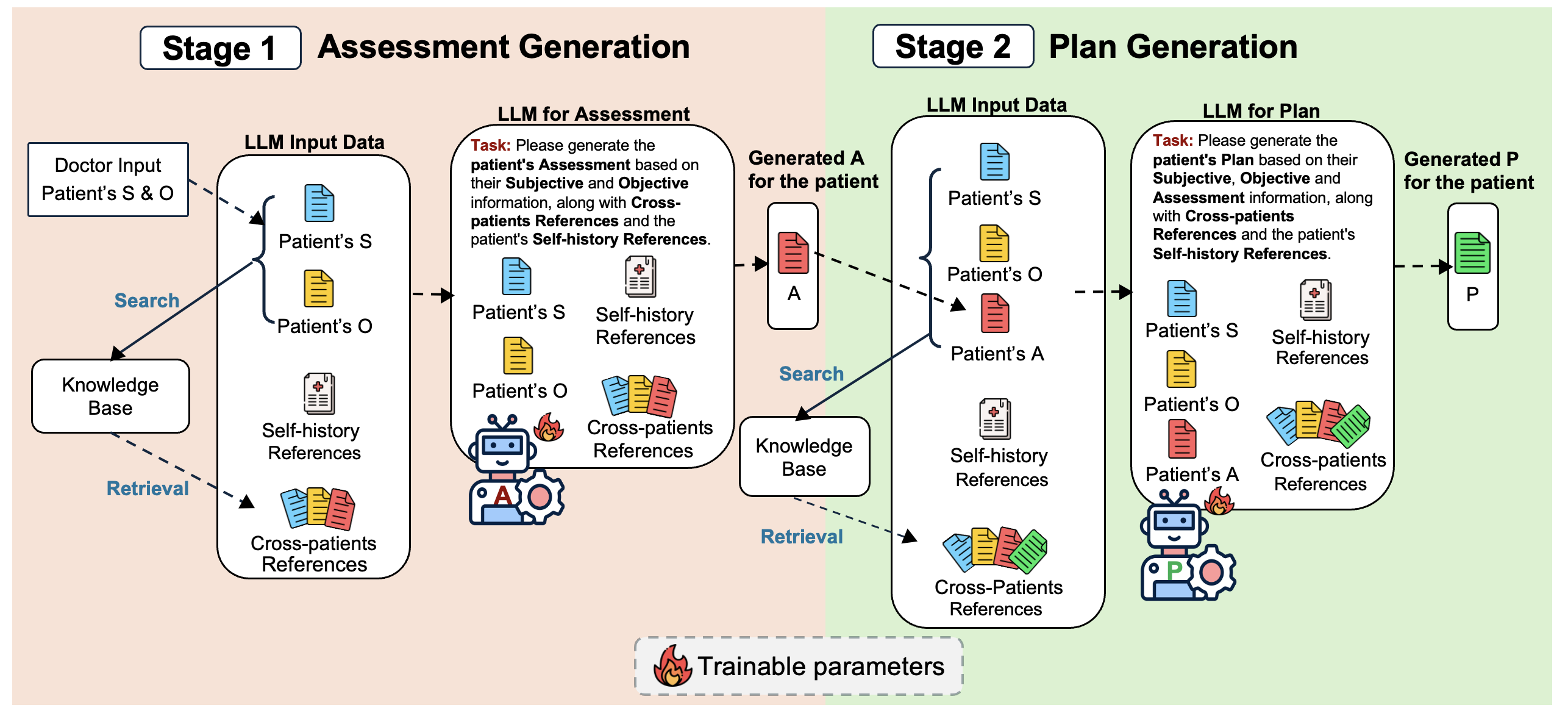}
    \caption{Overall architecture of the proposed {\ours} framework.}
    \label{fig:model}
\end{figure*}

To obtain accurate and personalized clinical plannings that align with physician workflow, we present \ours{}, a trustworthy clinical decision support system that employs a two-stage generation pipeline, mirroring the natural progression of clinical planning. To get high-quality planning, we propose to first generate an assessment based on the patient data, then create the treatment plan based on both the patient data and the generated assessment. This separation follows the established SOAP protocol, where clinicians first analyze symptoms and findings to form a diagnosis before determining appropriate interventions. We also explicitly separate S and O components in our prompts (see Appendix~\ref{appendix:prompt}), allowing the model to distinctly process patient-reported symptoms versus clinical observations---a key distinction that enhances clinical relevance. To enhance the personalization and accuracy of the generated plannings, we further leverage two types of references during generation: (1) self-history references---the patient's previous SOAP records, and (2) cross-patient references---similar cases from other patients. Specifically, for the $i$-th patient, we retrieve their latest $N_{\text{hist}}$ SOAP records as self-history references, formulated as $\mathcal{R}^{\text{hist}}{i}= { (S{j},O_{j},A_{j},P_{j}) \mid j \in {1,2,...,N_{\text{hist}}} }$.
Furthermore, to better align with the clinical reasoning patterns, we incorporate instruction tuning on the models that generate A and P before deploying our two-stage pipeline.
Figure~\ref{fig:model} illustrates the overall architecture of our inference workflow.

\subsection{Assessment Generation Stage}

In the Assessment Generation Stage, we integrated the patient's current $S$ and $O$ information with both self-history references $\mathcal{R}_{\text{hist}}$ and cross-patient references $\mathcal{R}^{SOA} = \{ (S_j, O_j, A_j) \}_{j=1}^{N_{\text{ref}}}$. To identify the most relevant cross-patient references, we employ a two-step retrieval process. First, we retrieve $N_{\text{sim}}$ candidate references $\mathcal{R}^{\text{SOA}}_{\text{sim}}$ via hybrid retrieval~\cite{Ma2020HybridFR,Bruch_2023,hsu2025datdynamicalphatuning} combining BM25~\cite{robertson1995okapi} and bi-encoder semantic search~\cite{karpukhin-etal-2020-dense}, leveraging both keyword matching and semantic similarity. Then, we refined this selection using a more computationally intensive but more accurate cross-encoder re-ranking model~\cite{nogueira2020passagererankingbert} that evaluates the fine-grained clinical relevance by jointly encoding the query and each candidate:
$$
\mathcal{R}^{SOA} = \text{Top-}N_{\text{ref}} \Big( \text{ReR}\big(\{S, O\}, \mathcal{R}^{\text{SOA}}_{\text{sim}} \big) \Big),
$$ where $\text{ReR}(\{S, O\}, \mathcal{R}^{\text{SOA}}_{\text{sim}})$ represents the cross-encoder re-ranking function that scores each reference in $\mathcal{R}^{\text{SOA}}_{\text{sim}}$ based on its relevance to the current case $\{S, O\}$. After obtaining the refined references, we combine the current $(S,O)$ with both $\mathcal{R}_{SOA}$ and $\mathcal{R}_{\text{hist}}$ to generate the assessment: $$
 A_{\text{gen}}=f_{\theta_A}(S,O,\mathcal{R}^{SOA}, \mathcal{R}^{\text{hist}}),
$$ where $A_{\text{gen}}$ denotes the generated assessment and $f_{\theta_A}$ represents the medical language model for assessment generation.

\subsection{Plan Generation Stage}

In the Plan Generation Stage, we utilized the generated assessment $A_{\text{gen}}$ along with the original $S$ and $O$ to retrieve and generate an appropriate treatment plan. Mirroring the clinical practice where physicians formulate treatment plans based on their diagnostic assessment and patient information, we employed another retrieval process to find relevant plan references $\mathcal{R}^{SOAP} = \{ (S_j, O_j, A_j, P_j) \}_{j=1}^{N_{\text{ref}}}$. Similar to the previous stage, we use a two-step retrieval approach. First, we retrieve $N_{\text{sim}}$ candidate references $\mathcal{R}^{\text{SOAP}}_{\text{sim}}$ via hybrid retrieval combining BM25 and bi-encoder semantic search. Then, we refined this selection using a cross-encoder re-ranking model:

\begin{footnotesize}
$$\mathcal{R}^{SOAP} = \text{Top-}N_{\text{ref}} \Big( \text{ReR}\big(\{S, O, A_{\text{gen}}\},\mathcal{R}^{\text{SOAP}}_{\text{sim}} \big) \Big),$$
\end{footnotesize}
where $\text{ReR}(\{S, O, A_{\text{gen}}\}, \mathcal{R}^{\text{SOAP}}_{\text{sim}})$ represents the cross-encoder re-ranking function that evaluates each reference in $\mathcal{R}^{\text{SOAP}}_{\text{sim}}$ based on its relevance to the current case with the generated assessment. After obtaining the refined references, we combined the current $(S,O,A_{\text{gen}})$ with both $\mathcal{R}^{SOAP}$ and $\mathcal{R}^{\text{hist}}$ to generate the treatment plan:
\begin{footnotesize}
$$P_{\text{gen}}=f_{\theta_P}(S,O,A_{\text{gen}},\mathcal{R}_{SOAP}, \mathcal{R}_{\text{hist}}),$$
\end{footnotesize} where $P_{\text{gen}}$ denotes the generated plan and $f_{\theta_P}$ represents the medical language model for plan generation.

\subsection{Information Alignment}
To align the models with the clinical reasoning pattern of our dataset, we instruction-tuned both the assessment generation model and plan generation model using the following objectives:
\begin{footnotesize}
$$\theta_A = \underset{\theta}{\text{argmin}} \sum_{i=1}^{N} \mathcal{L}(f_{\theta}(S_i, O_i, \mathcal{R}^{SOA}_{i}, \mathcal{R}^{\text{hist}}_{i}), A_i),$$
\end{footnotesize}
\begin{footnotesize}
$$\theta_P = \underset{\theta}{\text{argmin}} \sum_{i=1}^{N} \mathcal{L}(f_{\theta}(S_i, O_i, A_i,\mathcal{R}^{SOAP}_{i}, \mathcal{R}^{\text{hist}}_{i}), P_i),$$
\end{footnotesize} where $\mathcal{L}$ is the loss function, $N$ is the number of training samples, and $A_i$ and $P_i$ are the ground truth assessment and plan, respectively. This training process ensures that our models can properly interpret and utilize the medical context specific to our dataset.

\section{Experiments}

\subsection{Datasets}

This study utilized 350,684 outpatient and emergency EHR SOAP notes from 55,890 patients collected at Far Eastern Memorial Hospital (FEMH) in 2021. All data were de-identified prior to analysis. We preprocessed all SOAP notes by removing records shorter than two characters and normalizing text (eliminating newlines, redundant spaces, and consecutive punctuation). 

Unlike disease-specific approaches, our dataset encompasses general cases, ensuring broader applicability across clinical scenarios. To achieve this, we selected patients with three or more visits and employed a patient-centric sampling strategy. Specifically, records from 6,000 patients constituted our RAG knowledge base embedding, while an additional 3,000 randomly selected patient records were allocated into training and testing sets.

\subsection{Metrics}
For evaluation metrics, we used BLEU~\cite{papineni2002bleu}, METEOR~\cite{banerjee2005meteor}, ROUGE~\cite{lin2004rouge}, and BERTScore~\cite{zhang2019bertscore} using an independent inference script. Lexical similarity is evaluated using METEOR (Metric for Evaluation of Translation with Explicit Ordering) and BLEU (Bilingual Evaluation Understudy), with METEOR considering stemming and synonyms. ROUGE, which is the abbreviation of Recall-Oriented Understudy for Gisting Evaluation scores, compares the produced and reference summaries for the longest common subsequence (ROUGE-L) and n-gram overlaps (ROUGE-1, ROUGE-2). In order to properly evaluate text coherence and meaning, BERTScore balances recall and accuracy by using contextual embeddings to estimate semantic similarity beyond precise matches.

\begin{table*}[htbp]
\centering
\caption{Performance Comparison of Different Models and Settings for Plan Generation}
\resizebox{\textwidth}{!}{
\begin{tabular}{clccccccccc}
\hline
\textbf{Planning Method} & \textbf{Model} & \textbf{Self-history} & \textbf{Instruction Tuning} & \textbf{Cross-patient} & \textbf{BLEU} $\uparrow$ & \textbf{METEOR} $\uparrow$ & \textbf{ROUGE1} $\uparrow$ & \textbf{ROUGE2} $\uparrow$ & \textbf{ROUGE\_L $\uparrow$} & \textbf{Bertscore\_F1} $\uparrow$ \\ \hline
\multirow{14}{*}{\textbf{S+O$\rightarrow$P}} & o1 & \checkmark &  &  & 0.016399 & 0.140358 & 0.125431 & 0.046444 & 0.107900 & 0.817148 \\ \cline{2-11} 
 & GPT-4o & \checkmark &  &  & 0.028817 & 0.166348 & 0.154136 & 0.070183 & 0.139563 & 0.827025 \\ \cline{2-11} 
 & \multirow{4}{*}{Medical-Llama3-8B} &  & \checkmark &  & 0.052796 & 0.173414 & 0.220035 & 0.129617 & 0.214548 & 0.847451 \\
 &  & \checkmark &  &  & 0.178594 & 0.306591 & 0.343440 & 0.274914 & 0.340154 & 0.867276 \\
 &  & \checkmark & \checkmark &  & 0.291157 & 0.477312 & 0.535286 & 0.434203 & 0.531056 & 0.907823 \\
 &  & \checkmark & \checkmark & \checkmark & 0.307380 & 0.501418 & 0.559243 & 0.456576 & 0.554414 & 0.911653 \\ \cline{2-11} 
 & \multirow{4}{*}{Bio-Medical-Llama3-8B} &  & \checkmark &  & 0.061325 & 0.188050 & 0.235100 & 0.148139 & 0.228682 & 0.850004 \\
 &  & \checkmark &  &  & 0.112796 & 0.217000 & 0.235758 & 0.174116 & 0.230855 & 0.848391 \\
 &  & \checkmark & \checkmark &  & 0.299377 & 0.486631 & 0.544217 & 0.441678 & 0.539558 & 0.908784 \\
 &  & \checkmark & \checkmark & \checkmark & 0.309457 & 0.501485 & 0.557870 & 0.456750 & 0.553876 & 0.911572 \\ \cline{2-11} 
 & \multirow{4}{*}{Medical-Mixtral-7B-v2k} &  & \checkmark &  & 0.067164 & 0.196569 & 0.249694 & 0.156125 & 0.243456 & 0.852184 \\
 &  & \checkmark &  &  & 0.170338 & 0.311579 & 0.365305 & 0.285245 & 0.360484 & 0.869952 \\
 &  & \checkmark & \checkmark &  & 0.298256 & 0.482994 & 0.541785 & 0.442677 & 0.537791 & 0.910507 \\
 &  & \checkmark & \checkmark & \checkmark & 0.312393 & 0.510814 & 0.570339 & 0.464942 & 0.565761 & 0.914185 \\ \hline
\multirow{3}{*}{\textbf{\begin{tabular}[c]{@{}c@{}}S+O$\rightarrow$A$\rightarrow$P \\ (\ours{})\end{tabular}}} & \textbf{Bio-Medical-Llama3-8B} & \checkmark & \checkmark & \checkmark & \textbf{0.312238} & \textbf{0.516716} & \textbf{0.574780} & \textbf{0.467528} & \textbf{0.569738} & \textbf{0.915024} \\
 & \textbf{Medical-Llama3-8B} & \checkmark & \checkmark & \checkmark & \textbf{0.314718} & \textbf{0.516189} & \textbf{0.576113} & \textbf{0.469581} & \textbf{0.571199} & \textbf{0.915500} \\
 & \textbf{Medical-Mixtral-7B-v2k} & \checkmark & \checkmark & \checkmark & \textbf{0.318286} & \textbf{0.521312} & \textbf{0.581657} & \textbf{0.475762} & \textbf{0.577055} & \textbf{0.917194} \\ \hline
\end{tabular}}
\label{tab:results}
\end{table*}

\subsection{Implementation Details}

We utilized prompt engineering techniques and applied LoRA for parameter-efficient fine-tuning. Specifically, we instruction-tuned several open-source LLMs—Medical-Llama3-8B~\cite{Medical-Llama-3-8B-we-used}, Medical-Mixtral-7B-v2k~\cite{Medical-Mixtral-7B-v2k}, and Bio-Medical-Llama3-8B~\cite{ContactDoctor_Bio-Medical-Llama-3-8B}—using the Unsloth framework~\cite{unsloth}. To support long-context retrieval in our RAG-based design, we adopted OpenAI’s text-embedding-3-large model~\cite{openai-embedding} for semantic similarity search, and used VoyageAI Reranker-2~\cite{voyageai-rerank} as a cross-encoder model to re-rank the retrieved candidates. For baseline comparison, we additionally evaluated two general-purpose models: o1~\cite{openai-o1} and GPT-4o~\cite{openai-gpt-4o}, without domain-specific adaptation.

We set $N_{\text{hist}} = 20$ and $N_{\text{ref}} = 10$ for our RAG module, retrieving $N_{\text{sim}} = 80$ initial candidates based on semantic similarity. To evaluate \ours{}, we simulated clinical diagnostic processes by using the first $N{-}2$ visits as history $\mathcal{R}_{\text{hist}}$ and the second-to-last visit as the training target for patients with $N$ visits, while the first $N{-}1$ visits and the most recent visit were used as history and evaluation target respectively during testing. We conducted ablation experiments with various configurations by selectively enabling components in our pipeline, including: \textbf{Self-history}, \textbf{Instruction Tuning}, \textbf{Cross-patient References}, \textbf{Direct Plan Generation}, and a \textbf{Two-step Approach with Pre-plan Assessment}. Additional implementation details, including training environment and hyperparameter settings, are provided in Appendix~\ref{appendix:model_details}.

\begin{table*}[!h]
    \centering
    \caption{Comparison Performance in Patient-Specific Assessments Generation}
    \resizebox{\textwidth}{!}{
    \begin{tabular}{lccccccccc}
\hline
\textbf{Model} & \textbf{Self-history} & \textbf{Instruction Tuning} & \textbf{Cross-patient} & \textbf{BLEU $\uparrow$} & \textbf{METEOR $\uparrow$} & \textbf{ROUGE1 $\uparrow$} & \textbf{ROUGE2 $\uparrow$} & \textbf{ROUGE\_L $\uparrow$} & \textbf{Bertscore\_F1 $\uparrow$} \\ \hline
\multirow{3}{*}{Medical-Mixtral-7B-v2k}    & \checkmark       &                   &              & 0.302052      & 0.469219        & 0.535851        & 0.437234        & 0.532359          & 0.905538               \\
                          & \checkmark       & \checkmark        &              & 0.484695      & 0.653686        & 0.704872        & 0.606026        & 0.700879          & 0.940547               \\
                          & \checkmark       & \checkmark        & \checkmark   & \textbf{0.493051}      & \textbf{0.665725}        & \textbf{0.715743}        & \textbf{0.616415}        & \textbf{0.712651}          & \textbf{0.942709}               \\ \hline
\multirow{3}{*}{Bio-Medical-Llama3-8B}     & \checkmark       &                   &              & 0.234989      & 0.35864         & 0.378168        & 0.310427        & 0.372989          & 0.872104               \\
                          & \checkmark       & \checkmark        &              & 0.479665      & 0.645509        & 0.697491        & 0.596622        & 0.693297          & 0.938073               \\
                          & \checkmark       & \checkmark        & \checkmark   & \textbf{0.490539}      & \textbf{0.664329}        & \textbf{0.717387}        & \textbf{0.61274}         & \textbf{0.713025}          & \textbf{0.942353}               \\ \hline
\multirow{3}{*}{Medical-Llama3-8B}         & \checkmark       &                   &              & 0.303517      & 0.431265        & 0.466276        & 0.401507        & 0.463519          & 0.889349               \\
                          & \checkmark       & \checkmark        &              & 0.474254      & 0.641288        & 0.692784        & 0.594512        & 0.68923           & 0.937197               \\
                          & \checkmark       & \checkmark        & \checkmark   & \textbf{0.487554}      & \textbf{0.658435}        & \textbf{0.713324}        & \textbf{0.610607}        & \textbf{0.710027}          & \textbf{0.941513}               \\ \hline
\end{tabular}}
    \label{tab:table3}
\end{table*}

\begin{figure*}[h]
    \centering
    \includegraphics[width=0.8\textwidth]{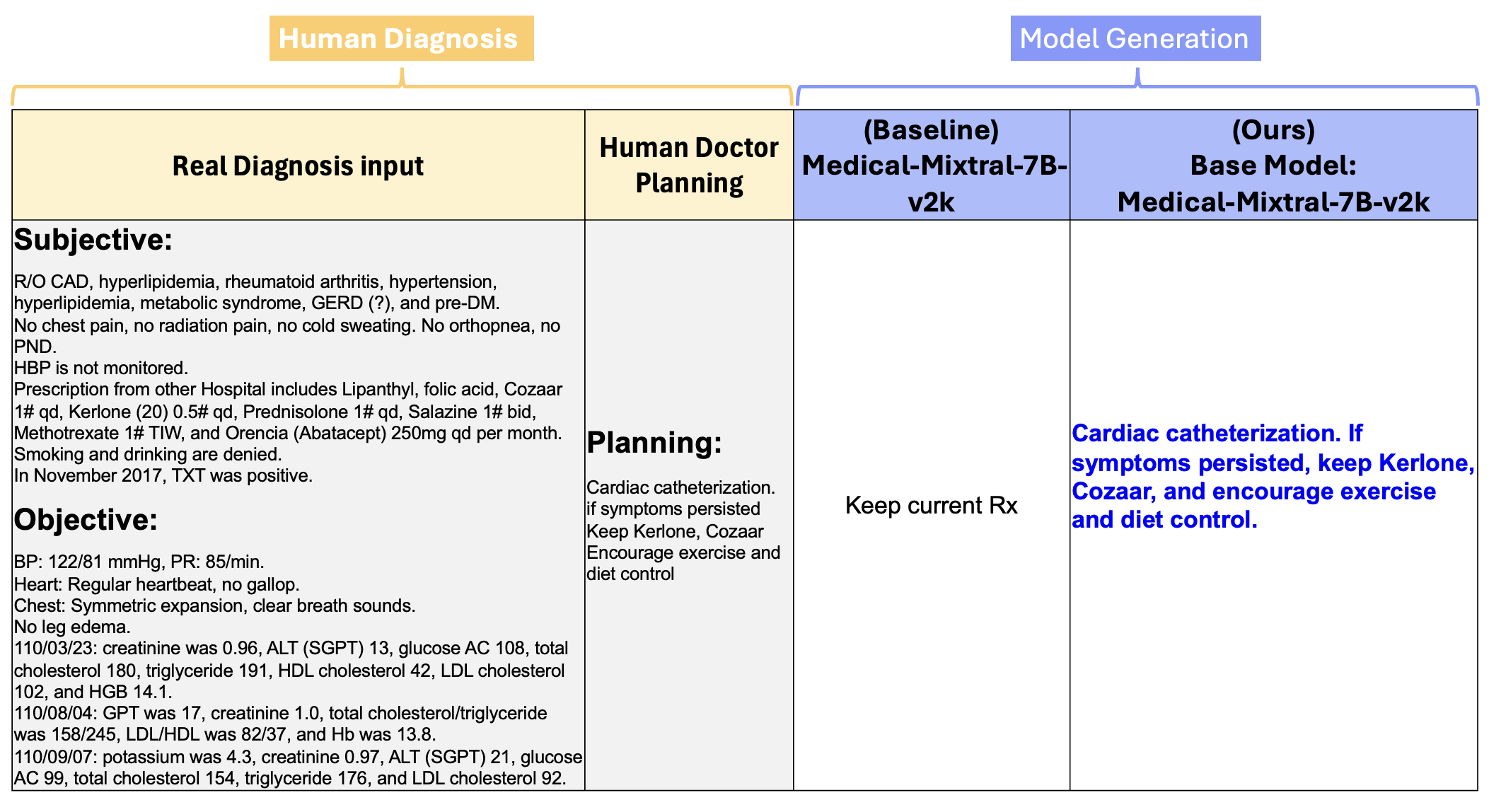}
    \caption{Plan Generation Results: Human Doctor, Baseline LLM, and \ours{}}
    \label{fig:samples_main}
\end{figure*}

\subsection{Results}

\textbf{\indent Does \ours{} help improve clinical planning?} In Table~\ref{tab:results}, our SOAP-inspired \ours{} (S+O→A→P) outperforms the baseline approach (S+O→P) across all backbone models and evaluation metrics. For example, on the Medical-Llama3-8B model, \ours{} increases BLEU from 0.307 to 0.315 and METEOR from 0.501 to 0.516. 
This is likely because \ours{} structures LLM reasoning in a manner that mirrors real-world clinical workflows, leading to more reliable planning.

\textbf{Does \ours{} help improve clinical assessment?} In Table~\ref{tab:table3}, \ours{} method integrates historical cross-patient assessments records, and consistently promotes base versions of all backbones on all metrics. In particular, on the Medical-Llama3-8B backbone, \ours{} improves METEOR by 2\%, with ROUGE1 and ROUGE2 by 2\% and 1.5\%, respectively. Similar gains are also observed in other models. This improvement likely results from the inference-time knowledge augmentation provided by the cross-patient information, which enriches the contextual input and helps the model generate more accurate and trustworthy assessments.



\textbf{How do we better support personalized planning?} As shown in Table~\ref{tab:results}, integrating patient history and cross-patient information via RAG enables our \ours{} to significantly enhance plan generation across all evaluated models. For instance, adding RAG in the instruction-tuned Medical-Llama3-8B model raises BLEU from 0.052 to 0.307 and METEOR from 0.173 to 0.501.  This might due to the enriched contextual input brought by the RAG, which augments the knowledge in the inference time and help the model to generate more trustworthy clinical plans.

\textbf{How do our generated treatment plans compare qualitatively to baseline approaches?}
Figure \ref{fig:samples_main} illustrates the qualitative improvement in clinical decision support capabilities. When presented with a complex patient case featuring multiple cardiovascular risk factors (hyperlipidemia, hypertension, metabolic syndrome, and pre-diabetes), the baseline Medical-Mixtral-7B-v2k model produced only a simplistic "Keep current Rx" recommendation—missing critical diagnostic and treatment components necessary for evidence-based care. In contrast, our approach generated a comprehensive clinical recommendation: "Cardiac catheterization. If symptoms persist, keep Kerlone, Cozaar, and encourage exercise and diet control." This output demonstrates enhanced capabilities to: (1) prioritize appropriate diagnostic procedures, (2) implement condition-based medication management, and (3) incorporate preventive lifestyle interventions for modifiable risk factors.

\begin{figure}
    \centering
    \includegraphics[width=0.5\textwidth]{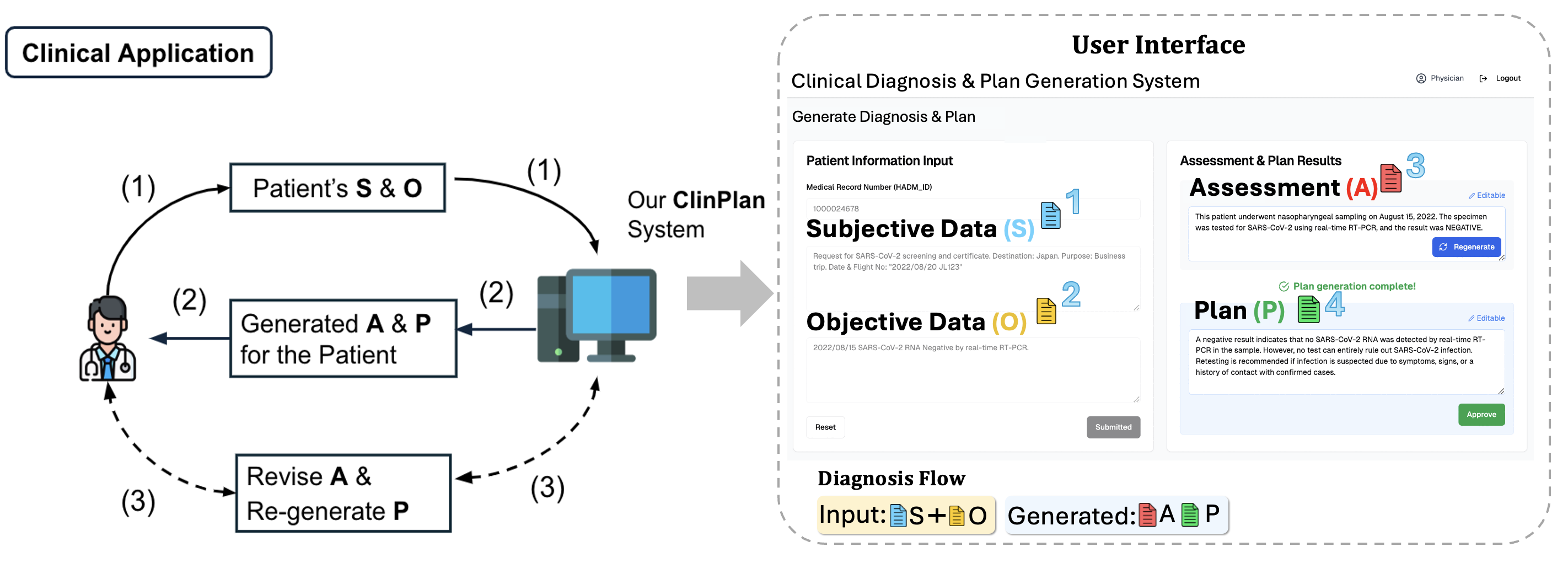}
    \caption{Overview of the Clinical Application of the \ours{} System}
    \label{fig:interface}
\end{figure}

\section{Clinical Application Demo and System Design}

To demonstrate the real-world applicability of our Plan generation system, we developed a clinical prototype that has been reviewed by practicing physicians for viability in actual healthcare settings. An overview of the clinical interface is shown in Figure~\ref{fig:interface}. Our system works as follows: The physician first inputs the patient's S and O, and the system generates A and P based on these inputs. At the same time, physicians can modify A according to their clinical judgment and regenerate P, while our system can update retrievals through RAG, which leverages a knowledge base of patient SOAP notes. The more specific technical architecture of the backend system is shown in Figure~\ref{fig:model}. The frontend is developed using React, the backend is based on FastAPI service, and communication between frontend and backend is conducted through RESTful API. The core of the system includes two specialized LLMs, responsible for generating A and P respectively. The system uses Microsoft SQL (MSSQL) database to store patient historical data, and enhances semantic retrieval and case matching through vector embedding using Weaviate database. 

The detailed system architecture is provided in Appendix~\ref{appendix:sys_arch}.

\section{Conclusion}

In this study, we introduced {\ours}, a novel approach leveraging LLMs with RAG to produce personalized treatment plans following the SOAP methodology. By structuring LLM reasoning into a two-stage process mirroring physician workflows, {\ours} generates assessments before formulating plans informed by patient-specific context. Empirical evaluation on an in-house dataset demonstrated promising outcomes and potential for future LLM diagnostic generation research work.


\bibliography{references}

\appendix



\section{System Architecture}\label{appendix:sys_arch}

\begin{figure*}[h]
    \centering
    \includegraphics[trim={0 2cm 0 3cm},clip,width=1\textwidth]{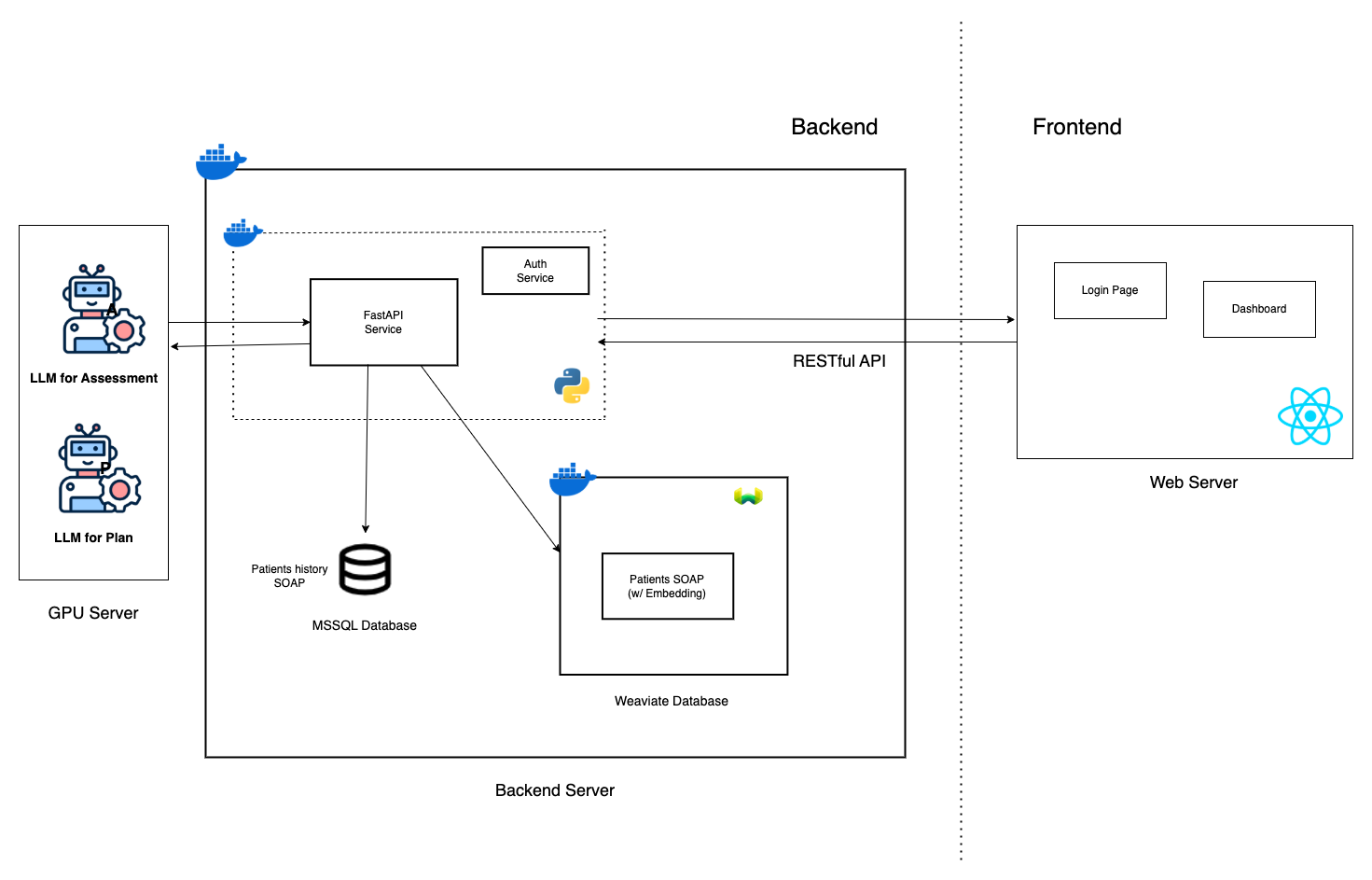}
    \caption{\ours{} System Architecture.}
    \label{fig:system_arch}
\end{figure*}

Our system architecture is designed for real-world deployment, ensuring robustness and efficiency when handling large-scale requests in the future. As illustrated in Figure \ref{fig:system_arch}, the backend is implemented using FastAPI, designed for high concurrency and efficient request handling. Instead of synchronous API calls, which may lead to memory overload or timeouts, we adopt an asynchronous task management approach. Upon receiving input, the backend assigns a unique task ID and forwards the request to the LLM. Once processing is completed, the system returns the results alongside the task ID, ensuring a seamless experience without blocking other requests.

\ours{} integrates two databases to support its functionality. Microsoft SQL Server stores structured patient data, allowing efficient retrieval of the latest consultation records using MRN (Medical Record Number) as a key. Additionally, Weaviate, a vector database, is employed to store a large repository of past patient records. These enable retrieval-augmented generation (RAG), allowing the system to identify cross-patient similar cases and provide physicians with relevant contextual information.

The user interface is developed using React, providing an intuitive web-based platform for physicians to interact with the system. The underlying LLM is deployed on our GPU server, which is equipped with NVIDIA hardware, ensuring efficient real-time inference and responsiveness.

\subsection{Implementation Details}\label{appendix:model_details}

We instruction-tuned three domain-specific LLMs—Medical-Llama3-8B~\cite{Medical-Llama-3-8B-we-used}, Medical-Mixtral-7B-v2k~\cite{Medical-Mixtral-7B-v2k}, and Bio-Medical-Llama3-8B~\cite{ContactDoctor_Bio-Medical-Llama-3-8B}—using the Unsloth framework~\cite{unsloth} for efficient adaptation with long-context support. All models were trained on NVIDIA RTX 6000 Ada Generation GPUs with Low-Rank Adaptation (LoRA), dynamically adjusted for each model's architecture. A maximum sequence length of 65{,}536 tokens was used to accommodate extended patient histories and cross-patient references. The training employed the AdamW optimizer in 8-bit precision, along with a cosine learning rate scheduler and a warm-up phase equal to 1.6\% of the total steps.

For semantic retrieval, we used OpenAI's text-embedding-3-large model~\cite{openai-embedding}, which supports high-dimensional dense representations suitable for medical content. As our cross-encoder model, we employed the VoyageAI Reranker-2~\cite{voyageai-rerank}, which was used to re-rank the semantically retrieved candidates in our RAG pipeline. All experiments were conducted under consistent hardware and software configurations to ensure comparability.


\section{Generation Samples}
Figure \ref{fig:samples_main} demonstrates a significant improvement in clinical decision support capabilities between the best baseline Medical-Mixtral-7B-v2k model and \ours{} with the Medical-Mixtral-7B-v2k model as the base model. The baseline model only produced the simple result, ``Keep current Rx'', while dealing with a complicated patient scenario that included several cardiovascular risk factors, such as hyperlipidemia, hypertension, metabolic syndrome, and pre-diabetes. This result indicates a troubling missing core diagnostic and treatment components necessary for evidence-based treatment.

In contrast, our approach produced a comprehensive, clinically sound recommendation that aligns remarkably with expert human physician judgment. Our model's output ``Cardiac catheterization. If symptoms persist, keep Kerlone, Cozaar, and encourage exercise and diet control'' demonstrates the model's enhanced capacity to (1) prioritize appropriate diagnostic procedures for suspected coronary artery disease, (2) implement condition-based medication management strategies, and (3) incorporate preventive lifestyle interventions addressing modifiable risk factors.

When a subset of the generated samples was presented to physicians at Far Eastern Memorial Hospital (FEMH) for evaluation, the proposed method demonstrated approximately 66\% improvement in clinical assessments compared to the baseline approach.

These findings highlight how combining RAG with two-stage targeted instruction tuning of LLMs can substantially improve AI clinical reasoning capabilities, potentially enhancing model utility in real-world medical decision support systems. Our proposed approach exhibits precise clinical reasoning, addressing both urgent diagnostic needs and long-term illness management concerns, suggesting promising directions for medical AI applications in healthcare settings.


\section{Prompt Template}\label{appendix:prompt}

\begin{figure*}[h]
    \centering
    \includegraphics[width=\textwidth]{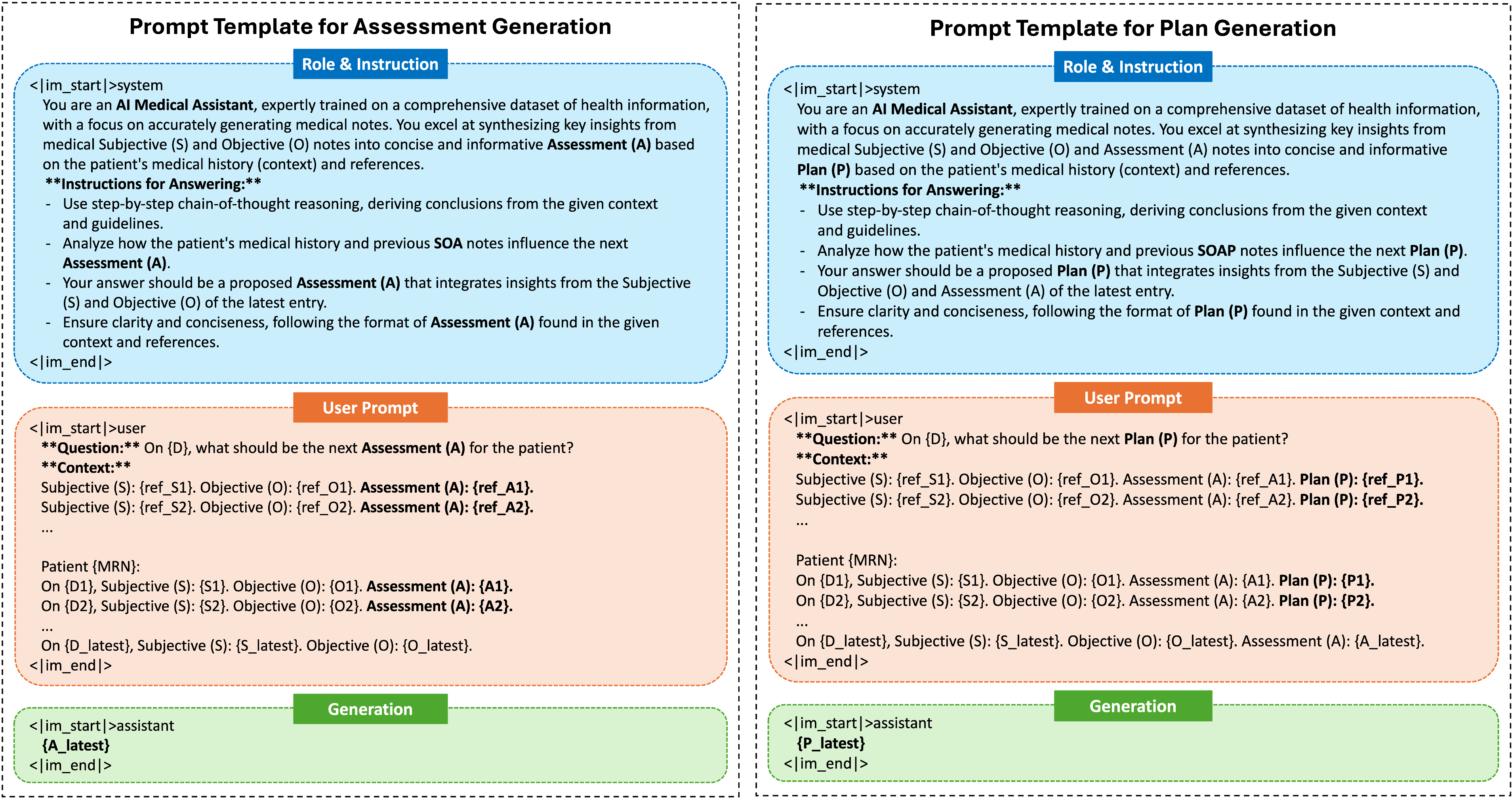}
    \caption{Prompt Template for Generation}
    \label{fig:prompt}
\end{figure*}

We present our prompt template (Figure \ref{fig:prompt}) to guide the generation by the LLMs. The left figure outlines the Assessment Generation template, while the right figure introduces the Plan Generation template. Each template contains three key sections:

\begin{itemize}
    \item \textbf{Role \& Instruction}: Directs an AI Medical Assistant to synthesize patient data using chain-of-thought reasoning.
    \item \textbf{User Prompt}: Provides structured query formats with placeholders for patient-specific information.
    \item \textbf{Generation}: Designates space for AI-generated content ([A\_latest] or [P\_latest]).
\end{itemize}

\section{Limitation}
The main limitation of this study lies in the data source and applicability. Our models are trained on EHR SOAP records from a specific hospital, which may limit its generalizability to other medical institutions or specialties. Additionally, while \ours{} employs retrieval-augmented generation (RAG) to enhance accuracy, it is still subject to inherent biases in language models, potentially leading to generating content that does not fully align with medical standards. These limitations highlight the need for continuous improvements and rigorous evaluation in real-world settings.

\end{document}